# Bethe-ADMM for Tree Decomposition based Parallel MAP Inference

**Qiang Fu Huahua Wang Arindam Banerjee**
Department of Computer Science & Engineering
University of Minnesota, Twin Cities
Minneapolis, MN 55455
{qifu, huwang, banerjee}@cs.umn.edu

## Abstract

We consider the problem of maximum a posteriori (MAP) inference in discrete graphical models. We present a parallel MAP inference algorithm called Bethe-ADMM based on two ideas: tree-decomposition of the graph and the alternating direction method of multipliers (ADMM). However, unlike the standard ADMM, we use an inexact ADMM augmented with a Bethe-divergence based proximal function, which makes each subproblem in ADMM easy to solve in parallel using the sum-product algorithm. We rigorously prove global convergence of Bethe-ADMM. The proposed algorithm is extensively evaluated on both synthetic and real datasets to illustrate its effectiveness. Further, the parallel Bethe-ADMM is shown to scale almost linearly with increasing number of cores.

## 1 Introduction

Given a discrete graphical model with known structure and parameters, the problem of finding the most likely configuration of the states is known as the *maximum a posteriori* (MAP) inference problem [23]. Existing approaches to solving MAP inference problems on graphs with cycles often consider a graph-based linear programming (LP) relaxation of the integer program [4, 18, 22] .

To solve the graph-based LP relaxation problem, two main classes of algorithms have been proposed. The first class of algorithms are dual LP algorithms [7, 9, 11, 19, 20, 21], which uses the dual decomposition and solves the dual problem. The two main approaches to solving the dual problems are block coordinate descent [7] and sub-gradient algorithms [11]. The coordinate descent algorithms are empirically faster, however, they may not reach the dual optimum since the dual problem is not strictly convex. Recent advances in coordinate descent algorithms perform tree-block updates [20, 21]. The sub-gradient methods, which are guaranteed to converge to the global optimum, can be slow in practice. For a detailed discussion on dual MAP algorithms, we refer the readers to [19]. The second class of algorithms are primal LP algorithms like the proximal algorithm [18]. The advantage of such algorithms is that it can choose different Bregman divergences as proximal functions which can take the graph structure into account. However, the proximal algorithms do not have a closed form update at each iteration in general and thus lead to double-loop algorithms.

As solving MAP inference in large scale graphical models is becoming increasingly important, in recent work, parallel MAP inference algorithms [14, 15] based on the alternating direction method of multipliers (ADMM) [1] have been proposed. As a primal-dual algorithm, ADMM combines the advantage of dual decomposition and the method of multipliers, which is guaranteed to converge globally and at a rate of $O(1/T)$ even for non-smooth problems [24]. ADMM has also been successfully used to solve large scale problem in a distributed manner [1].

Design of efficient parallel algorithms based on ADMM by problem decomposition has to consider a key tradeoff between the number of subproblems and the size of each subproblem. Having several simple subproblems makes solving each problem easy, but one has to maintain numerous dual variables to achieve consensus. On the other hand, having a few subproblems makes the number of constraints small, but each subproblem needs an elaborate often iterative algorithm, yielding a double-loop. Existing ADMM based algorithms for MAP inference [14, 15] decompose the problem into several simple subproblems, often based on single edges or local factors, so that the subproblems are easy to solve. However, to enforce consensus among the shared variables, such methods have to use dual variables proportional to the number of edges or local factors, which can make convergence slow on large graphs.

To overcome the limitations of existing ADMM methods for MAP inference, we propose a novel parallel algorithm based on tree decomposition. The individual trees need not be spanning and thus includes both edge decompo-sition and spanning tree decomposition as special cases. Compared to edge decomposition, tree decomposition has the flexibility of increasing the size of subproblems and reducing the number of subproblems by considering the graph structure. Compared to the tree block coordinate descent [20], which works with one tree at a time, our algorithm updates all trees in parallel. Note that the tree block coordinate descent algorithm in [21] updates disjoint trees within a forest in parallel, whereas our updates consider overlapping trees in parallel.

However, tree decomposition raises a new problem: the subproblems cannot be solved efficiently in the ADMM framework and requires an iterative algorithm, yielding a double-loop algorithm [14, 18]. To efficiently solve the subproblem on a tree, we propose a novel inexact ADMM algorithm called Bethe-ADMM, which uses a Bregman divergence induced by Bethe entropy on a tree, instead of the standard quadratic divergence, as the proximal function. The resulting subproblems on each tree can be solved exactly in linear time using the sum-product algorithm [12]. However, the proof of convergence for the standard ADMM does not apply to Bethe-ADMM. We prove global convergence of Bethe-ADMM and establish a $O(1/T)$ convergence rate, which is the same as the standard ADMM [8, 24]. Overall, Bethe-ADMM overcomes the limitations of existing ADMM based MAP inference algorithms [14, 15] and provides the flexibility required in designing efficient parallel algorithm through: (i) Tree decomposition, which can take the graph structure into account and greatly reduce the number of variables participating in the consensus and (ii) the Bethe-ADMM algorithm, which yields efficient updates for each subproblem.

We compare the performance of Bethe-ADMM with existing methods on both synthetic and real datasets and illustrate four aspects. First, Bethe-ADMM is faster than existing primal LP methods in terms of convergence. Second, Bethe-ADMM is competitive with existing dual methods in terms of quality of solutions obtained. Third, in certain graphs, tree decomposition leads to faster convergence than edge decomposition for Bethe-ADMM. Forth, parallel Bethe-ADMM, based on Open MPI, gets substantial speed-ups over sequential Bethe-ADMM. In particular, we show almost linear speed-ups with increasing number of cores on a graph with several million nodes.

The rest of the paper is organized as follows: We review the MAP inference problem in Section 2. In Section 3, we introduce the Bethe-ADMM algorithm and prove its global convergence. We discuss empirical evaluation in Section 4, and conclude in Section 5.

# 2 Background and Related Work

We first introduce some basic background on Markov Random Fields (MRFs). Then we briefly review existing ADMM based MAP inference algorithms in the literature. We mainly focus on pairwise MRFs and the discussions can be easily carried over to MRFs with general factors.

## 2.1 Problem Definition

A pairwise MRF is defined on an undirected graph $G = (V, E)$, where $V$ is the vertex set and $E$ is the edge set. Each node $u \in V$ has a random variable $X_u$ associated with it, which can take value $x_u$ in some discrete space $\mathcal{X} = \{1, \ldots, k\}$. Concatenating all the random variables $X_u$, $\forall u \in V$, we obtain an $n$ dimensional random vector $\boldsymbol{X} = \{X_u | u \in V\} \in \mathcal{X}^n$. We assume that the distribution $P$ of $\boldsymbol{X}$ is a Markov Random Field [23], meaning that it factors according to the structure of the undirected graph $G$ as follows: With $f_u : \mathcal{X} \mapsto \mathbb{R}, \forall u \in V$ and $f_{uv} : \mathcal{X} \times \mathcal{X} \mapsto \mathbb{R}$, $\forall (u, v) \in E$ denoting nodewise and edgewise potential functions respectively, the distribution takes the form $P(\boldsymbol{x}) \propto \exp\left\{\sum_{u \in V} f_u(x_u) + \sum_{(u,v) \in E} f_{uv}(x_u, x_v)\right\}$.

An important problem in the context of MRF is that of *maximum a posteriori* (MAP) inference, which is the following integer programming (IP) problem:

$$\boldsymbol{x}^* \in \underset{\boldsymbol{x} \in \mathcal{X}^n}{\operatorname{argmax}} \left\{ \sum_{u \in V} f_u(x_u) + \sum_{(u,v) \in E} f_{uv}(x_u, x_v) \right\} . \quad (1)$$

The complexity of (1) depends critically on the structure of the underlying graph. When $G$ is a tree structured graph, the MAP inference problem can be solved efficiently via the max-product algorithm [12]. However, for an arbitrary graph $G$, the MAP inference algorithm is usually computationally intractable. The intractability motivates the development of algorithms to solve the MAP inference problem approximately. In this paper, we focus on the linear programming (LP) relaxation method [4, 22]. The LP relaxation of MAP inference problem is defined on a set of pseudomarginals $\mu_u$ and $\mu_{uv}$, which are non-negative, normalized and locally consistent [4, 22]:

$$\begin{aligned} \mu_u(x_u) \geq 0 \ , \quad & \forall u \in V \ , \\ \sum_{x_u \in \mathcal{X}_u} \mu_u(x_u) = 1, \quad & \forall u \in V \ , \\ \mu_{uv}(x_u, x_v) \geq 0, \quad & \forall (u, v) \in E \ , \\ \sum_{x_u \in \mathcal{X}_u} \mu_{uv}(x_u, x_v) = \mu_v(x_v), \quad & \forall (u, v) \in E \ . \end{aligned} \quad (2)$$

We denote the polytope defined by (2) as $L(G)$. The LP relaxation of MAP inference problem (1) becomes solving the following LP:

$$\max_{\boldsymbol{\mu} \in L(G)} \langle \boldsymbol{\mu}, \boldsymbol{f} \rangle \ . \quad (3)$$

If the solution $\boldsymbol{\mu}$ to (3) is an integer solution, it is guaranteed to be the optimal solution of (1). Otherwise, one can apply rounding schemes [17, 18] to round the fractional solution to an integer solution.

### 2.2 ADMM based MAP Inference Algorithms

In recent years, ADMM [14, 15] has been used to solve large scale MAP inference problems. To solve (3) using ADMM, we need to split nodes or/and edges and introduce equality constraints to enforce consensus among the shared variables. The algorithm in [14] adopts edge decomposition and introduces equality constraints for shared nodes. Let $d_i$ be the degree of node $i$. The number of equality constraints in [14] is $O(\sum_{i=1}^{|V|} d_i k)$, which is approximately equal to $O(|E|k)$. For binary pairwise MRFs, the subproblems for the ADMM in [14] have closed-form solutions. For multi-valued MRFs, however, one has to first binarize the MRFs which introduces additional $|V|k$ variables for nodes and $2|E|k^2$ variables for edges. The binarization process increases the number of factors to $O(|V| + 2|E|k)$ and the complexity of solving each subproblem increases to $O(|E|k^2 \log k)$. We note that in a recent work [13], the active set method is employed to solve the quadratic problem for arbitrary factors. A generalized variant of [14] which does not require binarization is presented in [15]. We refer to this algorithm as Primal ADMM and use it as a baseline in Section 4. Although each subproblem in primal ADMM can be efficiently solved, the number of equality constraints and dual variables is $O(2|E|k + |E|k^2)$. In [15], ADMM is also used to solve the dual of (1). We refer to this algorithm as the Dual ADMM algorithm and use it as a baseline in Section 4. The dual ADMM works for multi-valued MRFs and has a linear time algorithm for each subproblem, but the number of equality constraint is $O(2|E|k + |E|k^2)$.

## 3 Algorithm and Analysis

We first show how to solve (3) using ADMM based on tree decomposition. The resulting algorithm can be a double-loop algorithm since some updates do not have closed form solutions. We then introduce the Bethe-ADMM algorithm where every subproblem can be solved exactly and efficiently, and analyze its convergence properties.

### 3.1 ADMM for MAP Inference

We first show how to decompose (3) into a series of subproblems. We can decompose the graph $G$ into overlapping subgraphs and rewrite the optimization problem with consensus constraints to enforce the pseudo-marginals on subgraphs (local variables) to agree with $\boldsymbol{\mu}$ (global variable). Throughout the paper, we focus on tree-structured decompositions. To be more specific, let $\mathbb{T} = \{(V_1, E_1), \ldots, (V_{|\mathbb{T}|}, E_{|\mathbb{T}|})\}$ be a collection of subgraphs of $G$ which satisfies two criteria: (i) Each subgraph $\tau = (V_\tau, E_\tau)$ is a tree-structured graph and (ii) Each node $u \in V$ and each edge $(u, v) \in E$ is included in at least one subgraph $\tau \in \mathbb{T}$. We also introduce local variable $\boldsymbol{m}_\tau \in L(\tau)$ which is the pseudomarginal [4, 22] defined on each subgraph $\tau$. We use $\boldsymbol{\theta}_\tau$ to denote the potentials on subgraph $\tau$. We denote $\boldsymbol{\mu}_\tau$ as the components of global variable $\boldsymbol{\mu}$ that belong to subgraph $\tau$. Note that since $\boldsymbol{\mu} \in L(G)$ and $\tau$ is a tree-structured subgraph of $G$, $\boldsymbol{\mu}_\tau$ always lies in $L(\tau)$. In the newly formulated optimization problem, we will impose consensus constraints for shared nodes and edges. For the ease of exposition, we simply use the equality constraint $\boldsymbol{\mu}_\tau = \boldsymbol{m}_\tau$ to enforce the consensus.

The new optimization problem we formulate based on graph decomposition is then as follows:

$$\min_{\boldsymbol{m}_\tau, \boldsymbol{\mu}} \quad \sum_{\tau=1}^{|\mathbb{T}|} \rho_\tau \langle \boldsymbol{m}_\tau, \boldsymbol{\theta}_\tau \rangle \tag{4}$$

$$\text{subject to} \quad \boldsymbol{m}_\tau - \boldsymbol{\mu}_\tau = 0, \quad \tau = 1, \ldots, |\mathbb{T}| \tag{5}$$

$$\boldsymbol{m}_\tau \in L(\tau), \quad \tau = 1, \ldots, |\mathbb{T}| \tag{6}$$

where $\rho_\tau$ is a positive constant associated with each subgraph. We use the consensus constraints (5) to make sure that the pseudomarginals agree with each other in the shared components across all the tree-structured subgraphs. Besides the consensus constraints, we also impose feasibility constraints (6), which guarantee that, for each subgraph, the local variable $\boldsymbol{m}_\tau$ lies in $L(\tau)$. When the constraints (5) and (6) are satisfied, the global variable $\boldsymbol{\mu}$ is guaranteed to lie in $L(G)$.

To make sure that problem (3) and (4)-(6) are equivalent, we also need to guarantee that

$$\min_{\boldsymbol{m}_\tau} \sum_{\tau=1}^{|\mathbb{T}|} \rho_\tau \langle \boldsymbol{m}_\tau, \boldsymbol{\theta}_\tau \rangle = \max_{\boldsymbol{\mu}} \langle \boldsymbol{\mu}, \boldsymbol{f} \rangle \ , \tag{7}$$

assuming the constraints (5) and (6) are satisfied. It is easy to verify that, as long as (7) is satisfied, the specific choice of $\rho_\tau$ and $\boldsymbol{\theta}_\tau$ do not change the problem. Let $\mathbf{1}[.]$ be a binary indicator function and $\boldsymbol{l} = -\boldsymbol{f}$. For any positive $\rho_\tau, \forall \tau \in \mathbb{T}$, e.g., $\rho_\tau = 1$, a simple approach to obtaining the potential $\boldsymbol{\theta}_\tau$ can be:

$$\theta_{\tau,u}(x_u) = \frac{l_u(x_u)}{\sum_{\tau'} \rho_{\tau'} \mathbf{1}[u \in V_{\tau'}]}, u \in V_\tau \ ,$$

$$\theta_{\tau,uv}(x_u, x_v) = \frac{l_{uv}(x_u, x_v)}{\sum_{\tau'} \rho_{\tau'} \mathbf{1}[(u,v) \in E_{\tau'}]}, (u, v) \in E(\tau) \ .$$

Let $\boldsymbol{\lambda}_\tau$ be the dual variable and $\beta > 0$ be the penalty parameter. The following updates constitute a single iteration of the ADMM [1]:

$$\boldsymbol{m}_\tau^{t+1} = \underset{\boldsymbol{m}_\tau \in L(\tau)}{\operatorname{argmin}} \langle \boldsymbol{m}_\tau, \rho_\tau \boldsymbol{\theta}_\tau + \boldsymbol{\lambda}_\tau^t \rangle + \frac{\beta}{2} ||\boldsymbol{m}_\tau - \boldsymbol{\mu}_\tau^t||_2^2 \ , \tag{8}$$

$$\boldsymbol{\mu}^{t+1} = \underset{\boldsymbol{\mu}}{\operatorname{argmin}} \sum_{\tau=1}^{|\mathbb{T}|} \left( -\langle \boldsymbol{\mu}_\tau, \boldsymbol{\lambda}_\tau^t \rangle + \frac{\beta}{2} ||\boldsymbol{m}_\tau^{t+1} - \boldsymbol{\mu}_\tau||_2^2 \right) , \tag{9}$$

$$\boldsymbol{\lambda}_\tau^{t+1} = \boldsymbol{\lambda}_\tau^t + \beta(\boldsymbol{m}_\tau^{t+1} - \boldsymbol{\mu}_\tau^{t+1}) \ . \tag{10}$$

In the tree based ADMM (8)-(10), the equality constraints are only required for shared nodes and edges. Assume there are $m$ shared nodes and the shared node $v_i$ has $C_i^v$ copies and there are $n$ shared edges and the shared edge $e_j$ has $C_j^e$ copies. The total number of equality constraints is $O(\sum_{i=1}^m C_i^v k + \sum_{j=1}^n C_j^e k^2)$. A special case of tree decomposition is edge decomposition, where only nodes are shared. In edge decomposition, $n = 0$ and the number of equality constraints is $O(\sum_{i=1}^m C_i^v k)$, which is approximately equal to $O(|E|k)$ and similar to [14]. In general, the number of shared nodes and edges in tree decomposition is much smaller than that in edge decomposition. The smaller number of equality constraints usually lead to faster convergence in achieving consensus. Now, the problem turns to whether the updates (8) and (9) can be solved efficiently, which we analyze below:

**Updating $\boldsymbol{\mu}$:** Since we have an unconstrained optimization problem (9) and the objective function decomposes component-wisely, taking the derivatives and setting them to zero yield the solution. In particular, let $S_u$ be the set of subgraphs which contain node $u$, for the node components, we have:

$$\mu_u^{t+1}(x_u) = \frac{1}{|S_u|\beta} \sum_{\tau \in S_u} \left(\beta m_{\tau,u}^{t+1}(x_u) + \lambda_{\tau,u}^t(x_u)\right) . \quad (11)$$

(11) can be further simplified by observing that $\sum_{\tau \in S_u} \lambda_{\tau,u}^t(x_u) = 0$ [1]:

$$\mu_u^{t+1}(x_u) = \frac{1}{|S_u|} \sum_{\tau=1}^{T} m_{\tau,u}^{t+1}(x_u) . \quad (12)$$

Let $S_{uv}$ be the subgraphs which contain edge $(u, v)$. The update for the edge components can be similarly derived as:

$$\mu_{u,v}^{t+1}(x_u, x_v) = \frac{1}{|S_{uv}|} \sum_{\tau \in S_{uv}} m_{\tau,uv}^{t+1}(x_u, x_v) . \quad (13)$$

**Updating $\boldsymbol{m}_\tau$:** For (8), we need to solve a quadratic optimization problem for each tree-structured subgraph. Unfortunately, we do not have a close-form solution for (8) in general. One possible approach, similar to the proximal algorithm, is to first obtain the solution $\tilde{\boldsymbol{m}}_\tau$ to the unconstrained problem of (8) and then project $\tilde{\boldsymbol{m}}_\tau$ to $L(\tau)$:

$$\boldsymbol{m}_\tau = \operatorname*{argmin}_{\boldsymbol{m} \in L(\tau)} ||\boldsymbol{m} - \tilde{\boldsymbol{m}}_\tau||_2^2 . \quad (14)$$

If we adopt the cyclic Bregman projection algorithm [2] to solve (14), the algorithm becomes a double-loop algorithm, i.e., the cyclic projection algorithm projects the solution to each individual constraint of $L(\tau)$ until convergence and the projection algorithm itself is iterative. We refer to this algorithm as the Exact ADMM and use it as a baseline in Section 4.

### 3.2 Bethe-ADMM

Instead of solving (8) exactly, a common way in inexact ADMMs [10, 25] is to linearize the objective function in (8), i.e., the first order Taylor expansion at $\boldsymbol{m}_\tau^t$, and add a new quadratic penalty term such that

$$\boldsymbol{m}_\tau^{t+1} = \operatorname*{argmin}_{\boldsymbol{m}_\tau \in L(\tau)} \langle \mathbf{y}_\tau^t, \boldsymbol{m}_\tau - \boldsymbol{m}_\tau^t \rangle + \frac{\alpha}{2} \|\boldsymbol{m}_\tau - \boldsymbol{m}_\tau^t\|_2^2 , \quad (15)$$

where $\alpha$ is a positive constant and

$$\mathbf{y}_\tau^t = \rho_\tau \boldsymbol{\theta}_\tau + \boldsymbol{\lambda}_\tau^t + \beta(\boldsymbol{m}_\tau^t - \boldsymbol{\mu}_\tau^t) . \quad (16)$$

However, as discussed in the previous section, the quadratic problem (15) is generally difficult for a tree-structured graph and thus the conventional inexact ADMM does not lead to an efficient update for $\boldsymbol{m}_\tau$. By taking the tree structure into account, we propose an inexact minimization of (8) augmented with a Bregman divergence induced by the Bethe entropy. We show that the resulting proximal problem can be solved exactly and efficiently using the sum-product algorithm [12]. We prove that the global convergence of the Bethe-ADMM algorithm in Section 3.3.

The basic idea in the new algorithm is that we replace the quadratic term in (15) with a Bregman-divergence term $d_\phi(\boldsymbol{m}_\tau || \boldsymbol{m}_\tau^t)$ such that

$$\boldsymbol{m}_\tau^{t+1} = \operatorname*{argmin}_{\boldsymbol{m}_\tau \in L(\tau)} \langle \mathbf{y}_\tau^t, \boldsymbol{m}_\tau - \boldsymbol{m}_\tau^t \rangle + \alpha d_\phi(\boldsymbol{m}_\tau || \boldsymbol{m}_\tau^t) , \quad (17)$$

is efficient to solve for any tree $\tau$. Expanding the Bregman divergence and removing the constants, we can rewrite (17) as

$$\boldsymbol{m}_\tau^{t+1} = \operatorname*{argmin}_{\boldsymbol{m}_\tau \in L(\tau)} \langle \mathbf{y}_\tau^t/\alpha - \nabla\phi(\boldsymbol{m}_\tau^t), \boldsymbol{m}_\tau \rangle + \phi(\boldsymbol{m}_\tau). \quad (18)$$

For a tree-structured problem, what convex function $\phi(\boldsymbol{m}_\tau)$ should we choose? Recall that $\boldsymbol{m}_\tau$ defines the marginal distributions of a tree-structured distribution $p_{\boldsymbol{m}_\tau}$ over the nodes and edges:

$$\boldsymbol{m}_{\tau,u}(x_u) = \sum_{\neg x_u} p_{\boldsymbol{m}_\tau}(x_1, \ldots, x_u, \ldots, x_n), \ \forall u \in V_\tau ,$$

$$\boldsymbol{m}_{\tau,uv}(x_u, x_v) = \sum_{\neg x_u, \neg x_v} p_{\boldsymbol{m}_\tau}(x_1, \ldots, x_u, x_v, \ldots, x_n), \ \forall (uv) \in E_\tau .$$

It is well known that the sum-product algorithm [12] efficiently computes the marginal distributions for a tree structured graph. It can also be shown that the sum-product algorithm solves the following optimization problem [23] for tree $\tau$ for some constant $\boldsymbol{\eta}_\tau$:

$$\max_{\boldsymbol{m}_\tau \in L(\tau)} \langle \boldsymbol{m}_\tau, \boldsymbol{\eta}_\tau \rangle + H_{Bethe}(\boldsymbol{m}_\tau) , \quad (19)$$

where $H_{Bethe}(\boldsymbol{m}_\tau)$ is the Bethe entropy of $\boldsymbol{m}_\tau$ defined as:

$$H_{Bethe}(\boldsymbol{m}_\tau) = \sum_{u \in V_\tau} H_u(\boldsymbol{m}_{\tau,u}) - \sum_{(u,v) \in E_\tau} I_{uv}(\boldsymbol{m}_{\tau,uv}) , \quad (20)$$

where $H_u(\boldsymbol{m}_{\tau,u})$ is the entropy function on each node $u \in V_\tau$ and $I_{uv}(\boldsymbol{m}_{\tau,uv})$ is the mutual information on each edge $(u,v) \in E_\tau$.

Combing (18) and (19), we set $\boldsymbol{\eta}_\tau = \nabla\phi(\boldsymbol{m}_\tau^t) - \mathbf{y}_\tau^t/\alpha$ and choose $\phi$ to be the negative Bethe entropy of $\boldsymbol{m}_\tau$ so that (18) can be solved efficiently in linear time via the sum-product algorithm.

For the sake of completeness, we summarize the Bethe-ADMM algorithm as follows :

$$\boldsymbol{m}_\tau^{t+1} = \underset{\boldsymbol{m}_\tau \in L(\tau)}{\operatorname{argmin}} \langle \mathbf{y}_\tau^t/\alpha - \nabla\phi(\boldsymbol{m}_\tau^t), \boldsymbol{m}_\tau\rangle + \phi(\boldsymbol{m}_\tau) \ , \quad (21)$$

$$\boldsymbol{\mu}^{t+1} = \underset{\boldsymbol{\mu}}{\operatorname{argmin}} \sum_{\tau=1}^{T} \left( -\langle \boldsymbol{\lambda}_\tau^t, \boldsymbol{\mu}_\tau\rangle + \frac{\beta}{2} ||\boldsymbol{m}_\tau^{t+1} - \boldsymbol{\mu}_\tau||_2^2 \right), \quad (22)$$

$$\boldsymbol{\lambda}_\tau^{t+1} = \boldsymbol{\lambda}_\tau^t + \beta(\boldsymbol{m}_\tau^{t+1} - \boldsymbol{\mu}_\tau^{t+1}) \ , \quad (23)$$

where $\mathbf{y}_\tau^t$ is defined in (16) and $-\phi$ is defined in (20).

### 3.3 Convergence

We prove the global convergence of the Bethe-ADMM algorithm. We first bound the Bregman divergence $d_\phi$:

**Lemma 1** *Let $\boldsymbol{\mu}_\tau$ and $\boldsymbol{\nu}_\tau$ be two concatenated vectors of the pseudomarginals on a tree $\tau$ with $n_\tau$ nodes. Let $d_\phi(\boldsymbol{\mu}_\tau||\boldsymbol{\nu}_\tau)$ be the Bregman divergence induced by the negative Bethe entropy $\phi$. Assuming $\alpha \geq \max_\tau\{\beta(2n_\tau - 1)^2\}$, we have*

$$\alpha d_\phi(\boldsymbol{\mu}_\tau||\boldsymbol{\nu}_\tau) \geq \frac{\beta}{2}\|\boldsymbol{\mu}_\tau - \boldsymbol{\nu}_\tau\|_2^2 \ . \quad (24)$$

*Proof:* Let $P_\tau(\boldsymbol{x})$ be a tree-structured distribution on a tree $\tau = (V_\tau, E_\tau)$, where $|V_\tau| = n_\tau$ and $|E_\tau| = n_\tau - 1$. The pseudomarginal $\boldsymbol{\mu}_\tau$ has a total of $2n_\tau - 1$ components, each being a marginal distribution. In particular, there are $n_\tau$ marginal distributions corresponding to each node $u \in V_\tau$, given by

$$\mu_{\tau,u}(x_u) = \sum_{\neg x_u} P_\tau(x_1, \ldots, x_u, \ldots, x_n) \ . \quad (25)$$

Thus, $\boldsymbol{\mu}_u$ is the marginal probability for node $u$.

Further, there are $n_\tau - 1$ marginal components corresponding to each edge $(u,v) \in E_\tau$, given by

$$\mu_{\tau,uv}(x_u, x_v) = \sum_{\neg(x_u,x_v)} P(x_1, \ldots, x_u, \ldots, x_v, \ldots, x_n) \ . \quad (26)$$

Thus, $\boldsymbol{\mu}_{uv}$ is the marginal probability for nodes $(u,v)$.

Let $\boldsymbol{\mu}_\tau, \boldsymbol{\nu}_\tau$ be two pseudomarginals defined on tree $\tau$ and $P_{\boldsymbol{\mu}_\tau}, P_{\boldsymbol{\nu}_\tau}$ be the corresponding tree-structured distributions. Making use of (25), we have

$$\|P_{\boldsymbol{\mu}_\tau} - P_{\boldsymbol{\nu}_\tau}\|_1 \geq \|\boldsymbol{\mu}_{\tau,u} - \boldsymbol{\nu}_{\tau,u}\|_1, \quad \forall u \in V_\tau \ . \quad (27)$$

Similarly, for each edge, we have the following inequality because of (26)

$$\|P_{\boldsymbol{\mu}_\tau} - P_{\boldsymbol{\nu}_\tau}\|_1 \geq \|\boldsymbol{\mu}_{\tau,uv} - \boldsymbol{\nu}_{\tau,uv}\|_1, \ \forall (u,v) \in E_\tau \ . \quad (28)$$

Adding them together gives

$$(2n_\tau - 1)\|P_{\boldsymbol{\mu}_\tau} - P_{\boldsymbol{\nu}_\tau}\|_1 \geq \|\boldsymbol{\mu}_\tau - \boldsymbol{\nu}_\tau\|_1 \geq \|\boldsymbol{\mu}_\tau - \boldsymbol{\nu}_\tau\|_2 \ . \quad (29)$$

According to Pinsker's inequality [3], we have

$$\begin{aligned} d_\phi(\boldsymbol{\mu}_\tau||\boldsymbol{\nu}_\tau) = KL(P_{\boldsymbol{\mu}_\tau}, P_{\boldsymbol{\nu}_\tau}) &\geq \frac{1}{2}\|P_{\boldsymbol{\mu}_\tau} - P_{\boldsymbol{\nu}_\tau}\|_1^2 \\ &\geq \frac{1}{2(2n_\tau - 1)^2}\|\boldsymbol{\mu}_\tau - \boldsymbol{\nu}_\tau\|_2^2 \ . \quad (30) \end{aligned}$$

Multiplying $\alpha$ on both sides and letting $\alpha \geq \beta(2n_\tau - 1)^2$ complete the proof. ■

To prove the convergence of the objective function, we define a residual term $R_\tau^{t+1}$ as

$$R_\tau^{t+1} = \rho_\tau \langle \boldsymbol{m}_\tau^{t+1} - \boldsymbol{\mu}_\tau^*, \boldsymbol{\theta}_\tau \rangle \ , \quad (31)$$

where $\boldsymbol{\mu}_\tau^*$ is the optimal solution for tree $\tau$. We show that $R_\tau^{t+1}$ satisfies the following inequality:

**Lemma 2** *Let $\{\boldsymbol{m}_\tau, \boldsymbol{\mu}_\tau, \boldsymbol{\lambda}_\tau\}$ be the sequences generated by Bethe-ADMM. Assume $\alpha \geq \max_\tau\{\beta(2n_\tau - 1)^2\}$. For any $\boldsymbol{\mu}_\tau^* \in L(\tau)$, we have*

$$\begin{aligned} R_\tau^{t+1} &\leq \langle \boldsymbol{\lambda}_\tau^t, \boldsymbol{\mu}_\tau^* - \boldsymbol{m}_\tau^{t+1}\rangle + \alpha\big(d_\phi(\boldsymbol{\mu}_\tau^*||\boldsymbol{m}_\tau^t) - d_\phi(\boldsymbol{\mu}_\tau^*||\boldsymbol{m}_\tau^{t+1})\big) \\ &+ \frac{\beta}{2}\left(\|\boldsymbol{\mu}_\tau^* - \boldsymbol{\mu}_\tau^t\|_2^2 - \|\boldsymbol{\mu}_\tau^* - \boldsymbol{m}_\tau^t\|_2^2 - \|\boldsymbol{m}_\tau^{t+1} - \boldsymbol{\mu}_\tau^t\|_2^2\right) \ , \quad (32) \end{aligned}$$

*where $R_\tau^{t+1}$ is defined in (31).*

*Proof:* Since $\boldsymbol{m}_\tau^{t+1}$ is the optimal solution for (21), for any $\boldsymbol{\mu}_\tau^* \in L(\tau)$, we have the following inequality:

$$\langle \mathbf{y}_\tau^t + \alpha(\nabla\phi(\boldsymbol{m}_\tau^{t+1}) - \nabla\phi(\boldsymbol{m}_\tau^t)), \boldsymbol{\mu}_\tau^* - \boldsymbol{m}_\tau^{t+1}\rangle \geq 0 \ . \quad (33)$$

Substituting (16) into (33) and rearranging the terms, we have

$$\begin{aligned} R_\tau^{t+1} &\leq \langle \boldsymbol{\lambda}_\tau^t, \boldsymbol{\mu}_\tau^* - \boldsymbol{m}_\tau^{t+1}\rangle + \beta\langle \boldsymbol{m}_\tau^t - \boldsymbol{\mu}_\tau^t, \boldsymbol{\mu}_\tau^* - \boldsymbol{m}_\tau^{t+1}\rangle \\ &+ \alpha\langle \nabla\phi(\boldsymbol{m}_\tau^{t+1}) - \nabla\phi(\boldsymbol{m}_\tau^t), \boldsymbol{\mu}_\tau^* - \boldsymbol{m}_\tau^{t+1}\rangle \ . \quad (34) \end{aligned}$$

The second term in the RHS of (34) is equivalent to

$$\begin{aligned} 2\langle \boldsymbol{m}_\tau^t - \boldsymbol{\mu}_\tau^t, \boldsymbol{\mu}_\tau^* - \boldsymbol{m}_\tau^{t+1}\rangle &= \|\boldsymbol{m}_\tau^t - \boldsymbol{m}_\tau^{t+1}\|_2^2 \\ &+ \|\boldsymbol{\mu}_\tau^* - \boldsymbol{\mu}_\tau^t\|_2^2 - \|\boldsymbol{\mu}_\tau^* - \boldsymbol{m}_\tau^t\|_2^2 - \|\boldsymbol{m}_\tau^{t+1} - \boldsymbol{\mu}_\tau^t\|_2^2. \quad (35) \end{aligned}$$

The third term in the RHS of (34) can be rewritten as

$$\begin{aligned} &\langle \nabla\phi(\boldsymbol{m}_\tau^{t+1}) - \nabla\phi(\boldsymbol{m}_\tau^t), \boldsymbol{\mu}_\tau^* - \boldsymbol{m}_\tau^{t+1}\rangle \\ &= d_\phi(\boldsymbol{\mu}_\tau^*||\boldsymbol{m}_\tau^t) - d_\phi(\boldsymbol{\mu}_\tau^*||\boldsymbol{m}_\tau^{t+1}) - d_\phi(\boldsymbol{m}_\tau^{t+1}||\boldsymbol{m}_\tau^t). \quad (36) \end{aligned}$$

Substituting (35) and (36) into (34) and using Lemma 1 complete the proof. ■

We next show that the first term in the RHS of (32) satisfies the following result:

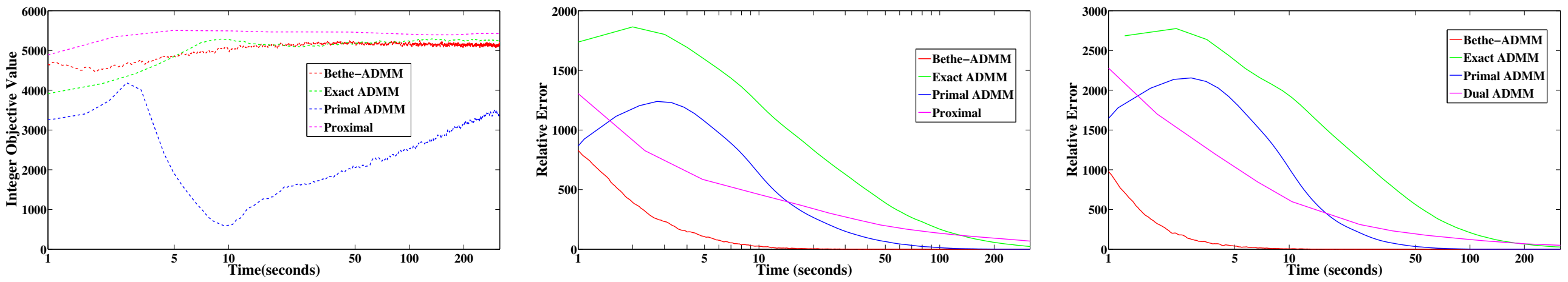

(a) Rounded solution with $a = 0.5$. (b) Relative error with $a = 0.5$. (c) Relative error with $a = 1$.

Figure 1: Results of Bethe-ADMM, Exact ADMM, Primal ADMM and proximal algorithms on two simulation datasets. Figure 1(a) plots the value of the decoded integer solution as a function of runtime (seconds). Figure 1(b) and 1(c) plot the relative error with respect to the optimal LP objective as a function of runtime (seconds). For Bethe-ADMM, we set $\alpha = \beta = 0.05$. For Exact ADMM, we set $\beta = 0.05$. For Primal ADMM, we set $\beta = 0.5$. Bethe-ADMM converges faster than other primal based algorithms.

**Lemma 3** *Let $\{\boldsymbol{m}_\tau, \boldsymbol{\mu}_\tau, \boldsymbol{\lambda}_\tau\}$ be the sequences generated by Bethe-ADMM. For any $\boldsymbol{\mu}_\tau^* \in L(\tau)$, we have*

$$\sum_{\tau=1}^{|\mathbb{T}|}\langle \boldsymbol{\lambda}_\tau^t, \boldsymbol{\mu}_\tau^* - \boldsymbol{m}_\tau^{t+1}\rangle \le \frac{1}{2\beta}(\|\boldsymbol{\lambda}_\tau^t\|_2^2 - \|\boldsymbol{\lambda}_\tau^{t+1}\|_2^2) + \frac{\beta}{2}\left(\|\boldsymbol{\mu}_\tau^* - \boldsymbol{m}_\tau^{t+1}\|_2^2 - \|\boldsymbol{\mu}_\tau^* - \boldsymbol{\mu}_\tau^{t+1}\|_2^2\right) .$$

*Proof:* Let $\mu_i$ be the $i$th component of $\boldsymbol{\mu}$. We augment $\boldsymbol{\mu}_\tau, \boldsymbol{m}_\tau$ and $\boldsymbol{\lambda}_\tau$ in the following way: If $\boldsymbol{\mu}_i$ is not a component of $\boldsymbol{\mu}_\tau$, we set $\boldsymbol{\mu}_{\tau,i} = 0, \boldsymbol{m}_{\tau,i} = 0$ and $\boldsymbol{\lambda}_{\tau,i} = 0$; otherwise, they are the corresponding components from $\boldsymbol{\mu}_\tau, \boldsymbol{m}_\tau$ and $\boldsymbol{\lambda}_\tau$ respectively. We can then rewrite (22) in the following equivalent component-wise form:

$$\mu_i^{t+1} = \underset{\mu_i}{\operatorname{argmin}} \sum_{\tau=1}^{|\mathbb{T}|}\left(\langle \lambda_{\tau,i}^t, m_{\tau,i}^{t+1} - \mu_{\tau,i}\rangle + \frac{\beta}{2}\|m_{\tau,i}^{t+1} - \mu_{\tau,i}\|_2^2\right) .$$

For any $\boldsymbol{\mu}_\tau^* \in L(\tau)$, we have the following optimality condition:

$$-\sum_{\tau=1}^{|\mathbb{T}|}\langle \lambda_{\tau,i}^t + \beta(m_{\tau,i}^{t+1} - \mu_{\tau,i}^{t+1}), \mu_{\tau,i}^* - \mu_{\tau,i}^{t+1}\rangle \ge 0 . \quad (37)$$

Combining all the components of $\boldsymbol{\mu}^{t+1}$, we can rewrite (37) in the following vector form:

$$-\sum_{\tau=1}^{|\mathbb{T}|}\langle \boldsymbol{\lambda}_\tau^t + \beta(\boldsymbol{m}_\tau^{t+1} - \boldsymbol{\mu}_\tau^{t+1}), \boldsymbol{\mu}_\tau^* - \boldsymbol{\mu}_\tau^{t+1}\rangle \ge 0 . \quad (38)$$

Rearranging the terms yields

$$\begin{aligned}
&\sum_{\tau=1}^{|\mathbb{T}|}\langle \boldsymbol{\lambda}_\tau^t, \boldsymbol{\mu}_\tau^* - \boldsymbol{m}_\tau^{t+1}\rangle \\
&\le \sum_{\tau=1}^{|\mathbb{T}|}\langle \boldsymbol{\lambda}_\tau^t, \boldsymbol{\mu}_\tau^{t+1} - \boldsymbol{m}_\tau^{t+1}\rangle - \sum_{\tau=1}^{|\mathbb{T}|}\beta\langle \boldsymbol{m}_\tau^{t+1} - \boldsymbol{\mu}_\tau^{t+1}, \boldsymbol{\mu}_\tau^* - \boldsymbol{\mu}_\tau^{t+1}\rangle \\
&= \sum_{\tau=1}^{|\mathbb{T}|}\langle \boldsymbol{\lambda}_\tau^t, \boldsymbol{\mu}_\tau^{t+1} - \boldsymbol{m}_\tau^{t+1}\rangle + \frac{\beta}{2}\sum_{\tau=1}^{|\mathbb{T}|}\left(\|\boldsymbol{\mu}_\tau^* - \boldsymbol{m}_\tau^{t+1}\|_2^2 - \|\boldsymbol{\mu}_\tau^* - \boldsymbol{\mu}_\tau^{t+1}\|_2^2 - \|\boldsymbol{\mu}_\tau^{t+1} - \boldsymbol{m}_\tau^{t+1}\|_2^2\right) . \quad (39)
\end{aligned}$$

Recall $\boldsymbol{\mu}_\tau^{t+1} - \boldsymbol{m}_\tau^{t+1} = \frac{1}{\beta}(\boldsymbol{\lambda}_\tau^t - \boldsymbol{\lambda}_\tau^{t+1})$ in (23), then

$$\langle \boldsymbol{\lambda}_\tau^t, \boldsymbol{\mu}_\tau^{t+1} - \boldsymbol{m}_\tau^{t+1}\rangle - \frac{\beta}{2}\|\boldsymbol{\mu}_\tau^{t+1} - \boldsymbol{m}_\tau^{t+1}\|_2^2 = \frac{1}{2\beta}(\|\boldsymbol{\lambda}_\tau^t\|_2^2 - \|\boldsymbol{\lambda}_\tau^{t+1}\|_2^2) . \quad (40)$$

Plugging (40) into (39) completes the proof. ■

We also need the following lemma which can be found in [6]. We omit the proof due to lack of space.

**Lemma 4** *Let $\{\boldsymbol{m}_\tau, \boldsymbol{\mu}_\tau, \boldsymbol{\lambda}_\tau\}$ be the sequences generated by Bethe-ADMM. Then*

$$\sum_{\tau=1}^{|\mathbb{T}|}\|\boldsymbol{m}_\tau^{t+1} - \boldsymbol{\mu}_\tau^t\|_2^2 \ge \sum_{\tau=1}^{|\mathbb{T}|}\|\boldsymbol{m}_\tau^{t+1} - \boldsymbol{\mu}_\tau^{t+1}\|_2^2 + \|\boldsymbol{\mu}_\tau^{t+1} - \boldsymbol{\mu}_\tau^t\|_2^2 .$$

**Theorem 1** *Assume the following hold: (1) $\boldsymbol{m}_\tau^0$ and $\boldsymbol{\mu}_\tau^0$ are uniform tree-structured distributions, $\forall \tau = 1, \ldots, |\mathbb{T}|$ (2) $\boldsymbol{\lambda}_\tau^0 = \boldsymbol{0}, \forall \tau = 1, \ldots, |\mathbb{T}|$; (3) $\max_\tau d_\phi(\boldsymbol{\mu}_\tau^* \| \boldsymbol{m}_\tau^0) = D_{\boldsymbol{\mu}}$; (4) $\alpha \ge \max_\tau\{\beta(2n_\tau - 1)^2\}$ holds. Denote $\bar{\boldsymbol{m}}_\tau^T = \frac{1}{T}\sum_{t=0}^{T-1}\boldsymbol{m}_\tau^t$ and $\bar{\boldsymbol{\mu}}_\tau^T = \frac{1}{T}\sum_{t=0}^{T-1}\boldsymbol{\mu}_\tau^t$. For any $T$ and the optimal solution $\boldsymbol{\mu}^*$, we have*

$$\sum_{\tau=1}^{|\mathbb{T}|}\left(\rho_\tau\langle \bar{\boldsymbol{m}}_\tau^T - \boldsymbol{\mu}_\tau^*, \theta_\tau\rangle + \frac{\beta}{2}\|\bar{\boldsymbol{m}}_\tau^T - \bar{\boldsymbol{\mu}}_\tau^T\|_2^2\right) \le \frac{D_{\boldsymbol{\mu}}\alpha|\mathbb{T}|}{T} .$$

*Proof:* Summing (32) over $\tau$ from 1 to $|\mathbb{T}|$ and using Lemma 3, we have:

$$\begin{aligned}
&\sum_{\tau=1}^{|\mathbb{T}|}\left(R_\tau^{t+1} + \frac{\beta}{2}\|\boldsymbol{m}_\tau^{t+1} - \boldsymbol{\mu}_\tau^t\|_2^2\right) \\
&\le \sum_{\tau=1}^{|\mathbb{T}|}\frac{1}{2\beta}(\|\boldsymbol{\lambda}_\tau^t\|_2^2 - \|\boldsymbol{\lambda}_\tau^{t+1}\|_2^2) + \frac{\beta}{2}(\|\boldsymbol{\mu}_\tau^* - \boldsymbol{\mu}_\tau^t\|_2^2 - \|\boldsymbol{\mu}_\tau^* - \boldsymbol{\mu}_\tau^{t+1}\|_2^2) \\
&+ \frac{\beta}{2}\left(\|\boldsymbol{\mu}_\tau^* - \boldsymbol{m}_\tau^{t+1}\|_2^2 - \|\boldsymbol{\mu}_\tau^* - \boldsymbol{m}_\tau^t\|_2^2\right) \\
&+ \alpha\left(d_\phi(\boldsymbol{\mu}_\tau^* \| \boldsymbol{m}_\tau^t) - d_\phi(\boldsymbol{\mu}_\tau^* \| \boldsymbol{m}_\tau^{t+1})\right) . \quad (41)
\end{aligned}$$

Summing over the above from $t = 0$ to $T-1$, we have

$$
\begin{aligned}
&\sum_{t=0}^{T-1}\sum_{\tau=1}^{|\mathbb{T}|}\left(R_\tau^{t+1} + \frac{\beta}{2}\|\boldsymbol{m}_\tau^{t+1} - \boldsymbol{\mu}_\tau^t\|_2^2\right) \\
&\le \sum_{\tau=1}^{|\mathbb{T}|}\frac{1}{2\beta}(\|\boldsymbol{\lambda}_\tau^0\|_2^2 - \|\boldsymbol{\lambda}_\tau^T\|_2^2) + \frac{\beta}{2}\left(\|\boldsymbol{\mu}_\tau^* - \boldsymbol{\mu}_\tau^0\|_2^2 - \|\boldsymbol{\mu}_\tau^* - \boldsymbol{\mu}_\tau^T\|_2^2\right) \\
&+ \frac{\beta}{2}\left(\|\boldsymbol{\mu}_\tau^* - \boldsymbol{m}_\tau^T\|_2^2 - \|\boldsymbol{\mu}_\tau^* - \boldsymbol{m}_\tau^0\|_2^2\right) \\
&+ \alpha\left(d_\phi(\boldsymbol{\mu}_\tau^*||\boldsymbol{m}_\tau^0) - d_\phi(\boldsymbol{\mu}_\tau^*||\boldsymbol{m}_\tau^T)\right) \\
&\le \sum_{\tau=1}^{|\mathbb{T}|}\frac{\beta}{2}\|\boldsymbol{\mu}_\tau^* - \boldsymbol{m}_\tau^T\|_2^2 + \alpha\left(d_\phi(\boldsymbol{\mu}_\tau^*||\boldsymbol{m}_\tau^0) - d_\phi(\boldsymbol{\mu}_\tau^*||\boldsymbol{m}_\tau^T)\right) \\
&\le \sum_{\tau=1}^{|\mathbb{T}|}\alpha d_\phi(\boldsymbol{\mu}_\tau^*||\boldsymbol{m}_\tau^0)\,, \qquad (42)
\end{aligned}
$$

where we use Lemma 1 to derive (42). Applying Lemma 4 and Jensen's inequality yield the desired bound. ■

Theorem 1 establishes the $O(1/T)$ convergence rate for the Bethe-ADMM in ergodic sense. As $T \to \infty$, the objective value $\sum_{\tau=1}^{|\mathbb{T}|}\rho_\tau\langle\bar{\boldsymbol{m}}_\tau^T, \theta_\tau\rangle$ converges to the optimal value and the equality constraints are also satisfied.

### 3.4 Extension to MRFs with General Factors

Although we present Bethe-ADMM in the context of pairwise MRFs, it can be easily generalized to handle MRFs with general factors. For a general MRF, we can view the dependency graph as a factor graph [12], a bipartite graph $G = (V \cup F, E)$, where $V$ and $F$ are disjoint set of variable nodes and factor nodes and $E$ is a set of edges, each connecting a variable node and a factor node. The distribution $P(\boldsymbol{x})$ takes the form: $P(\boldsymbol{x}) \propto \exp\left\{\sum_{u\in V} f_u(x_u) + \sum_{\alpha\in F} f_\alpha(\boldsymbol{x}_\alpha)\right\}$. The relaxed LP for general MRFs can be constructed in a similar fashion with that for pairwise MRFs.

We can then decompose the relaxed LP to subproblems defined on factor trees and impose equality constraints to enforce consistency on the shared variables among the subproblems. Each subproblem can be solved efficiently using the sum-product algorithm for factor trees and the Bethe-ADMM algorithm for general MRFs bears similar structure with that for pairwise MRFs.

## 4 Experimental Results

We compare the Bethe-ADMM algorithm with several other state-of-the-art MAP inference algorithms. We show the comparison results with primal based MAP inference algorithms in Section 4.1 and dual based MAP inference algorithm in Section 4.2 respectively. We also show in Section 4.3 how tree decomposition benefits the performance of Bethe-ADMM. We run experiments in Section 4.1-4.3 using sequential updates. To illustrate the scalability of our algorithm, we run parallel Bethe-ADMM on a multicore machine and show the linear speedup in Section 4.4.

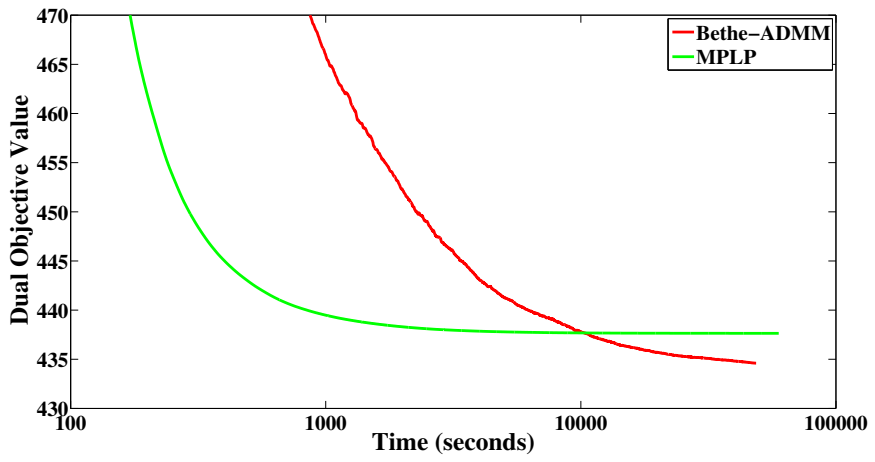

Figure 2: Both Bethe-ADMM and MPLP are run for sufficiently long, i.e., 50000 iterations. The dual objective value is plotted as a function of runtime (seconds). The MPLP algorithm gets stuck and does not reach the global optimum.

### 4.1 Comparison with Primal based Algorithms

We compare the Bethe-ADMM algorithm with the proximal algorithm [18], Exact ADMM algorithm and Primal ADMM algorithm [15]. For the proximal algorithm, we choose the Bregman divergence as the sum of KL-divergences across all node and edge distributions. Following the methodology in [18], we terminate the inner loop if the maximum constraint violation of $L(G)$ is less than $10^{-3}$ and set $w^t = t$. Similarly, in applying the Exact ADMM algorithm, we terminate the loop for solving $\boldsymbol{m}_\tau$ if the maximum constraint violation of $L(\tau)$ is less than $10^{-3}$. For the Exact ADMM and Bethe-ADMM algorithm, we use 'edge decomposition': each $\tau$ is simply an edge of the graph and $|\mathbb{T}| = |E|$. To obtain the integer solution, we use node-based rounding: $x_u^* = \operatorname{argmax}_{x_u} \mu_u(x_u)$.

We show experimental results on two synthetic datasets. The underlying graph of each dataset is a three dimensional $m\times n\times t$ grid. We generate the potentials as follows: We set the nodewise potentials as random numbers from $[-a, a]$, where $a > 0$. We set the edgewise potentials according to the Potts model, i.e., $\theta_{uv}(x_u, x_v) = b_{uv}$ if $x_u = x_v$ and 0 otherwise. We choose $b_{uv}$ randomly from $[-1, 1]$. The edgewise potentials penalize disagreement if $b_{uv} > 0$ and penalize agreement if $b_{uv} < 0$. We generate datasets using $m = 20, n = 20, t = 16, k = 6$ with varying $a$.

Figure 1(a) shows the plots of (1) on one synthetic dataset and we find that the algorithms have similar performances on other simulation datasets. We observe that all algorithms converge to the optimal value $\langle\boldsymbol{\mu}^*, \boldsymbol{f}\rangle$ of (3) and we plot the relative error with respect to the optimal value $|\langle\boldsymbol{\mu}^* - \boldsymbol{\mu}_t, \boldsymbol{f}\rangle|$ on the two datasets in Figure 1(b) and 1(c). Overall, the Bethe-ADMM algorithm converges faster than other primal algorithms. We observe that the proximal algorithm and Exact ADMM algorithm are the slowest, due to the sequential projection step. In terms of the decoded integer solution, the Bethe-ADMM, Exact ADMM and proximal algorithm have similar performances. We also note that a higher objective function value does not necessarily lead to a better decoded integer solution.

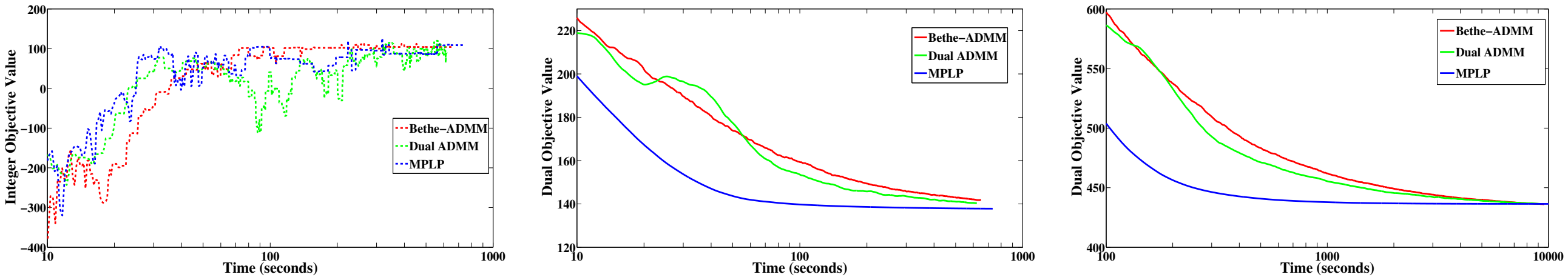

(a) Rounded integer solution on 1jo8. (b) Dual value on 1jo8. (c) Dual value on 1or7.

Figure 3: Results of Bethe-ADMM, MPLP and Dual ADMM algorithms on two protein design datasets. Figure 3(a) plots the the value of the decoded integer solution as a function of runtime (seconds). Figure 3(b) and 3(c) plot the dual value as a function of runtime (seconds). For Dual ADMM, we set $\beta = 0.05$. For Bethe-ADMM, we set $\alpha = \beta = 0.1$. Bethe-ADMM and Dual ADMM have similar performance in terms of convergence. All three methods have comparable performances for the decoded integer solution.

### 4.2 Comparison with Dual based Algorithms

In this section, we compare the Bethe-ADMM algorithm with the MPLP algorithm [7] and the Dual ADMM algorithm [15]. We conduct experiments on protein design problems [26]. In these problems, we are given a 3D structure and the goal is to find a sequence of amino-acids that is the most stable for that structure. The problems are modeled by nodewise and pairwise factors and can be posed as finding a MAP assignment for the given model. This is a demanding setting in which each problem may have hundreds of variables with 100 possible states on average.

We run the algorithms on two problems with different sizes [26], i.e., 1jo8 (58 nodes and 981 edges) and 1or7 (180 nodes and 3005 edges). For the MPLP and Dual ADMM algorithm, we plot the value of the integer programming problem (1) and its dual.. For Bethe-ADMM algorithm, we plot the value of dual LP of (3) and the integer programming problem (1). Note that although Bethe-ADMM and Dual ADMM have different duals, their optimal values are the same. We run the Bethe-ADMM based on edge decomposition. Figure 3 shows the result.

We observe that the MPLP algorithm usually converges faster, but since it is a coordinate ascent algorithm, it can stop prematurely and yield suboptimal solutions. Figure 2 shows that on the 1fpo dataset, the MPLP algorithm converges to a suboptimal solution. We note that the convergence time of the Bethe-ADM and Dual ADM are similar. The three algorithms have similar performance in terms of the decoded integer solution.

### 4.3 Edge based vs Tree based

In the previous experiments, we use 'edge decomposition' for the Bethe-ADMM algorithm. Since our algorithm can work for any tree-structured graph decomposition, we want to empirically study how the decomposition affects the performance of the Bethe-ADMM algorithm. In the following experiments, we show that if we can utilize the graph structure when decomposing the graph, the Bethe-ADMM algorithm will have better performance compared to simply using 'edge decomposition', which does not take the graph structure into account.

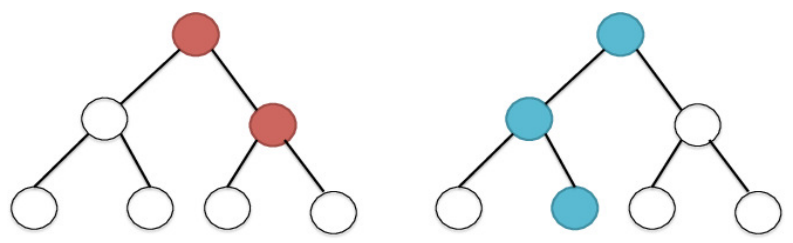

(a)

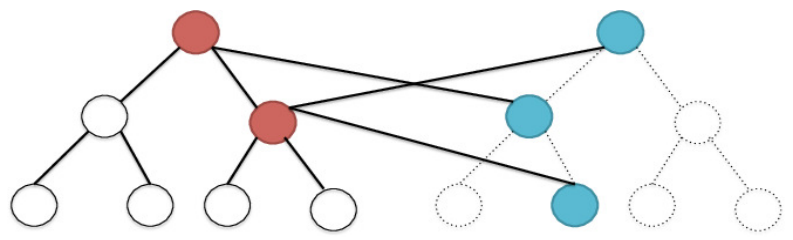

(b)

Figure 4: A simulation dataset with $m = 2$, $s = 7$ and $n = 3$. In 4(a), the red nodes ($S_{12}$) are sampled from tree 1 and the blue nodes ($D_{12}$) are sampled from tree 2. In 4(b) , sampled nodes are connected by cross-tree edges ($E_{12}$). Tree 1 with nodes in $D_{12}$ and edges in $E_{12}$ still form a tree, denoted by solid lines. This augmented tree is a tree-structured subgraph for Bethe-ADMM.

We conduct experiments on synthetic datasets. We generate MRFs whose dependency graphs consist of several tree-structured graphs and cross-tree edges to introduce cycles. To be more specific, we first generate $m$ binary tree structured MRFs each with $s$ nodes. Then for each ordered pair of tree-structured MRFs $(i, j), 1 \leq i, j \leq m, i \neq j$, we uniformly sample $n$ nodes from MRF $i$ with replacement and uniformly sample $n$ $(n \leq s)$ nodes from MRF $j$ without replacement, resulting in two node sets $S_{ij}$ and $D_{ij}$. We then connect the nodes in $S_{ij}$ and $D_{ij}$, denoting them as $E_{ij}$. We repeat this process for every pair of trees. By construction, the graph consisting of tree $i$, nodes in $D_{ij}$ and edges in $E_{ij}$, $\forall j \neq i$ is still a tree. We will use these $m$ augmented trees as the tree-structured subgraphs for the Bethe-ADMM algorithm. Figure 4 illustrates the graph generation and tree decomposition process. A simple calculation shows that for this particular tree decomposition, $O(m^2nk)$ equality constraints are maintained, while for edge decomposition, $O(msk + m^2nk)$ are maintained. When the graph has dominant tree structure, tree decomposition leads to much less number of equality constraints.

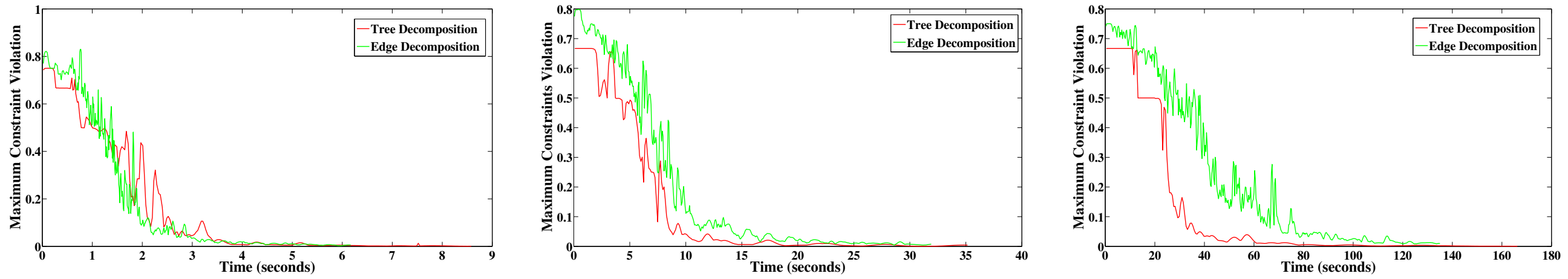

(a) $s = 1023$. (b) $s = 4095$. (c) $s = 16383$.

Figure 5: Results of Bethe-ADMM algorithms based on tree and edge decomposition on three simulation datasets with $m = 10, n = 20$. The maximum constraint violation in $L(G)$ is plotted as a function of runtime (seconds). For both algorithms, we set $\alpha = \beta = 0.05$. The tree based Bethe-ADMM algorithm has better performance than that of the edge based Bethe-ADMM when the tree structure is more dominant in $G$.

For the experiments, we run the Bethe-ADMM algorithm based on tree and edge decomposition with different values of $s$, keeping $m$ and $n$ fixed. It is easy to see that the tree structure becomes more dominant when $s$ becomes larger. Since we observe that both algorithms first converge to the optimal value of (3) and then the equality constraints are gradually satisfied, we evaluate the performance by computing the maximum constraint violation of $L(G)$ at each iteration for both algorithms. The faster the constraints are satisfied, the better the algorithm is. The results are shown in Figure 5. When the tree structure is not obvious, the two algorithms have similar performances. As we increase $s$ and the tree structure becomes more dominant, the difference between the two algorithms is more pronounced. We attribute the superior performance to the fact that for the tree decomposition case, much fewer number of equality constraints are imposed and each subproblem on tree can be solved efficiently using the sum-product algorithm.

### 4.4 Scalability Experiments on Multicores

The dataset used in this section is the Climate Research Unit (CRU) precipitation dataset [16], which has monthly precipitation from the years 1901-2006. The dataset is of high gridded spatial resolution ($360 \times 720$, i.e., 0.5 degree latitude $\times$ 0.5 degree longitude) and includes the precipitation over land.

Our goal is to detect major droughts based on precipitation. We formulate the problem as the one of estimating the most likely configuration of a binary MRF, where each node represents a location. The underlying graph is a three dimensional grid ($360 \times 720 \times 106$) with 7,146,520 nodes and each node can be in two possible states: dry and normal. We run the Bethe-ADMM algorithm on the CRU dataset and detect droughts based on the integer solution after node-based rounding. For the details of the this experiment, we refer to readers to [5]. Our algorithm successfully detects nearly all the major droughts of the last century. We also examine how the Bethe-ADMM algorithm scales on the CRU dataset with more than 7 million variables. We run the Open MPI code with different number of cores and the result in Figure 6 shows that we obtain almost linear speedup with the number of cores.

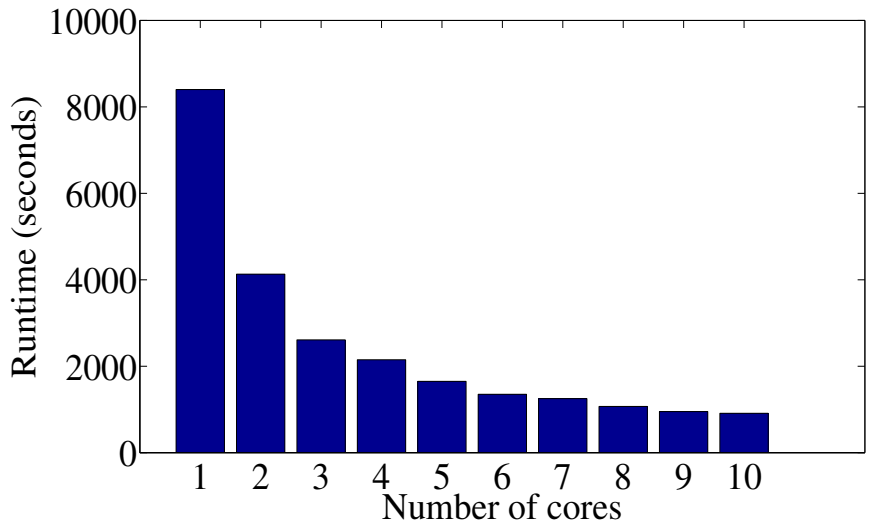

Figure 6: The Open MPI implementation of Bethe-ADMM has almost linear speedup on the CRU dataset with more than 7 million nodes.

## 5 Conclusions

We propose a provably convergent MAP inference algorithm for large scale MRFs. The algorithm is based on the 'tree decomposition' idea from the MAP inference literature and the alternating direction method from the optimization literature. Our algorithm solves the tree structured subproblems efficiently via the sum-product algorithm and is inherently parallel. The empirical results show that the new algorithm, in its sequential version, compares favorably to other existing approximate MAP inference algorithm in terms of running time and accuracy. The experimental results on large datasets demonstrate that the parallel version scales almost linearly with the number of cores in the multi-core setting.

## Acknowledgements

This research was supported in part by NSF CAREER Grant IIS-0953274, and NSF Grants IIS-1029711, IIS-0916750, and IIS-0812183. The authors would like to thank Stefan Liess and Peter K. Snyder for the helpful discussion on the CRU experiment. The authors are grateful for the technical support from the University of Minnesota Supercomputing Institute (MSI).